\documentclass[10pt,twocolumn,letterpaper]{article}

\usepackage{cvpr}              %
\usepackage{pdfpages}

\newcommand{\mparagraph}[1]{\vspace{1mm}\noindent{\textbf{#1.}\hspace{1mm}}}

\usepackage{import} %
\graphicspath{{figs/}}

\usepackage[ruled,noend]{algorithm2e} %

\SetAlFnt{\small}
\SetAlCapFnt{\small}
\SetAlCapNameFnt{\small}
\SetAlCapHSkip{0pt}
\IncMargin{-\parindent}
\usepackage{ifpdf}
\usepackage{graphicx}
\ifpdf
  
\else
\fi
\usepackage{color}
\definecolor{cyan}{cmyk}{1,0,0,0}
\definecolor{darkgreen}{rgb}{0,0.5,0}
\definecolor{orange}{rgb}{1,0.5,0}
\definecolor{magenta}{cmyk}{0,1,0,0}
\definecolor{darkyellow}{cmyk}{0,0,0.75,0}
\definecolor{gray}{rgb}{0.8,0.8,0.8}

\usepackage{amsmath} %

\usepackage[toc,page]{appendix} %
\usepackage{wrapfig} %
\usepackage{comment}
\usepackage{bm}
\usepackage{cuted} %
\usepackage{lipsum} %
\usepackage{mathtools} %
\usepackage{algorithmicx}
\usepackage{algpseudocode}
\usepackage{nicefrac}

\usepackage{gensymb}
\usepackage{float}
\usepackage{soul}
\usepackage{array}
\usepackage{multirow}
\usepackage{hhline}
\usepackage{setspace}
\usepackage{nicefrac}
\usepackage{makecell}

\algdef{SE}[DOWHILE]{Do}{doWhile}{\algorithmicdo}[1]{\algorithmicwhile\ #1}%

\makeatletter
\renewcommand{\ALG@beginalgorithmic}{\small}
\makeatother

\newcommand{\DELETE}[1]{} %
\newcommand{\IGNORE}[1]{}
\usepackage{datenumber}
\usepackage{calc}
\usepackage[mmddyyyy]{datetime}

\newcounter{datetoday}
\newcounter{diffyears}
\newcounter{diffmonths}
\newcounter{diffdays}

\newcommand{\difftoday}[3]{%
      \setmydatenumber{datetoday}{\the\year}{\the\month}{\the\day}%
      \setmydatenumber{diffdays}{#1}{#2}{#3}%
      \addtocounter{diffdays}{-\thedatetoday}%
      \ifnum\value{diffdays}>0
        \def\diffbefore{}%
        \def\diffafter{left}%
      \else
        \def\diffbefore{}%
        \def\diffafter{ago}%
        \setcounter{diffdays}{-\value{diffdays}}%
      \fi
      \setcounter{diffyears}{\value{diffdays}/365}%
      \setcounter{diffdays}{\value{diffdays}-365*\value{diffyears}}%
      \setcounter{diffmonths}{\value{diffdays}/30}%
      \setcounter{diffdays}{\value{diffdays}-30*\value{diffmonths}}%
      \diffbefore
      \ifnum\value{diffyears}=0
      \else
        \ifnum\value{diffyears}>1
            \thediffyears\space years,
        \else
            \thediffyears\space year,
        \fi
      \fi
      \ifnum\value{diffmonths}=0
      \else
        \ifnum\value{diffmonths}>1
            \thediffmonths\space months
        \else
            \thediffmonths\space month
        \fi
      \fi
      \ifnum\value{diffdays}=0
      \else
        \ifnum\value{diffdays}>1
            \thediffdays\space days
        \else
            \thediffdays\space day
        \fi
      \fi
      \diffafter
}

\definecolor{cvprblue}{rgb}{0.21,0.49,0.74}
\usepackage[pagebackref,breaklinks,colorlinks,allcolors=cvprblue]{hyperref}

\def\paperID{35997} %
\def\confName{CVPR}
\def\confYear{2026}

\title{Splat-Based Metal Artifact Reduction in Cone-Beam CT\\via Compact Attenuation Modeling}

\author{Kiseok Choi \hspace{1cm} Jaemin Cho \hspace{1cm} Inchul Kim \hspace{1cm} Min H. Kim\\[2mm]
KAIST\\[2mm]
{\tt\small \{kschoi;jmcho;ickim;mhkim\}@vclab.kaist.ac.kr}
}

\begin{document}
\maketitle
\begin{abstract}
X-ray computed tomography (CT) suffers from severe metal artifacts when high-attenuation objects such as dental fillings or orthopedic implants are present. 
These artifacts originate from the polychromatic nature of X-rays, where attenuation varies strongly with photon energy and material composition, breaking the monochromatic assumption used by conventional reconstruction algorithms. 
Recent neural rendering approaches attempt to address this mismatch through differentiable polychromatic projection models, 
but they still struggle with smoothness bias, loss of fine structures, and prohibitive computation when extended to large-scale cone-beam CT.
We introduce a splat-based metal artifact reduction framework that incorporates a physically grounded polychromatic forward model into a continuous Gaussian representation for cone-beam CT. 
Each Gaussian encodes the energy-dependent attenuation of the underlying material using a compact material parameterization, which enables efficient joint optimization of geometric and material properties without relying on a metal mask. 
This compact attenuation formulation captures the essential variation across biological tissues and metallic implants, allowing our model to explain metal-induced nonlinearity while preserving high-frequency structure.
Experiments on simulated and real cone-beam CT scans show that our method converges significantly faster and suppresses metal artifacts more effectively than existing reconstruction and neural field-based approaches.

\end{abstract}
    
\section{Introduction}
\label{sec:intro}

\begin{figure}[h]
	\centering
	\includegraphics[width=1.0\linewidth]{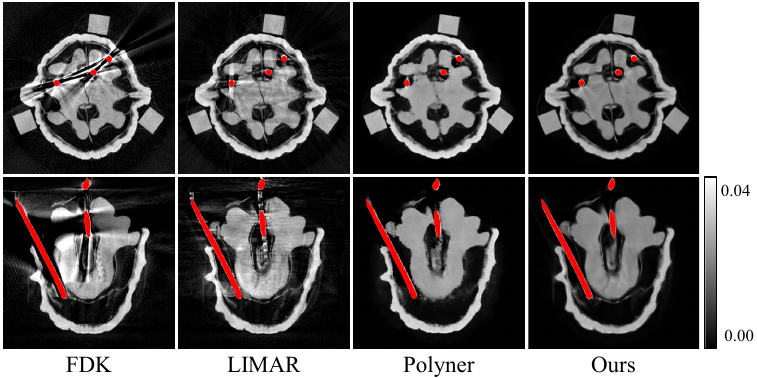}%
	\vspace{-2mm}
	\caption{\label{fig:teaser}
	Qualitative comparison of CBCT reconstructions for a real walnut specimen containing inserted metal pins (highlighted in red). 
	Each method is shown using a horizontal slice (top row) and a vertical slice (bottom row). 
	FDK~\cite{feldkamp1984practical}, when applied directly to the metal-inserted scan, produces strong beam hardening artifacts including dark streaks, severe shading, and distorted intensities. 
	LIMAR~\cite{limar} reduces some streaks but noticeably degrades overall structural fidelity. 
	Polyner~\cite{polyner} removes most streaks, however it introduces blurring and oversmoothing that wipe out fine details near metal boundaries. 
	In contrast, our method delivers the most faithful reconstruction, suppressing metal-induced artifacts while preserving crisp structural detail in both axial and sagittal views, demonstrating strong robustness under severe beam hardening in cone-beam CT.}
\vspace{-3mm}
\end{figure}

X-ray computed tomography (CT) is a non-destructive imaging technique that recovers the internal structure of an object by transmitting X-rays through it and measuring the attenuated energy. 
The geometry of CT systems has evolved from parallel-beam to fan-beam, and more recently to cone-beam configurations~\cite{kak_ct_book,kak2001principles}. 
Cone-beam CT (CBCT) enables volumetric reconstruction from a single rotation, making it widely used in medical diagnostics and industrial inspection.

Despite its efficiency, CBCT reconstruction is highly sensitive to inaccuracies in the underlying X-ray physics. 
In practice, the detected intensity follows a \emph{polychromatic} Beer–Lambert integration over the source spectrum and material-dependent attenuation coefficients. 
The attenuation of photons varies strongly with their energy and the material composition they traverse. Conventional reconstruction algorithms, including Feldkamp–Davis–Kress (FDK)  and analytical back-projection methods, assume \emph{monochromatic} attenuation, an assumption that breaks down in the presence of high-attenuation metals. 
This mismatch produces nonlinear distortions in the projection data, resulting in severe streaking and shading artifacts around metallic regions. 
The problem becomes especially pronounced in medical applications, where dental fillings or orthopedic implants introduce extreme beam hardening~\cite{hokamp2020quantification,boas2012ct,selles2024advances}.

Many approaches have attempted to mitigate metal artifacts in fan-beam CT~\cite{limar,nmar,polyner,yonsei}. 
Early methods corrected corrupted projection bins or interpolated missing regions, sometimes combined with image-domain priors~\cite{limar,nmar}. 
Although simple, these heuristics are fragile, geometry dependent, and unable to explain the nonlinear physics induced by metal. 
More recent work has shifted toward model-based reconstruction using differentiable polychromatic forward projections. 
Neural field-based methods draw inspiration from NeRFs~\cite{nerf} and optimize a continuous attenuation function to match polychromatic measurements~\cite{polyner,yonsei}. 
However, such representations often exhibit noise amplification, 
and their computational cost becomes prohibitive in full 3D cone-beam geometry.
Gaussian splatting offers a more efficient and expressive alternative for CBCT reconstruction, enabling faster optimization and improved structural detail~\cite{r2_gaussian}. However, existing splat-based methods are still limited to monochromatic attenuation models.

In this work, we introduce a splat-based metal artifact reduction framework that incorporates not only a differentiable polychromatic forward projection model into a continuous Gaussian representation, but also a differentiable reconstruction formulation that enables effective image-domain regularization. A key component of our method is a compact representation of energy-dependent material attenuation that captures essential variations across biological tissues and metallic implants using a physically grounded, low-dimensional model. This formulation enables efficient joint optimization of geometry and material properties, improving optimization stability, accelerating convergence, and effectively reducing metal-induced artifacts.
Our main contributions are as follows:
\begin{itemize}
\item We introduce a Gaussian-splatting-based CBCT reconstruction framework that incorporates both a physically grounded polychromatic forward-projection model and a differentiable reconstruction pipeline to achieve effective metal artifact reduction.
\item We propose a compact representation of energy-dependent attenuation that maintains physical realism while reducing the dimensionality of material optimization.
\item We provide a 3D CBCT dataset with realistic metal-induced artifacts and plan to release both the dataset and code upon publication.
\end{itemize}

\section{Related Work}
\label{sec:related_work}
\mparagraph{CBCT reconstruction}
Cone-beam CT reconstruction has been a central topic in X-ray imaging since the mid-20th century. The FDK algorithm~\cite{feldkamp1984practical}, a 3D extension of filtered back-projection based on the Fourier slice theorem, remains the industrial standard due to its simplicity and computational efficiency. 
However, FDK relies on a monochromatic attenuation assumption and uniform sampling, which makes it sensitive to interpolation errors and geometric inconsistencies, resulting in radial artifacts and limited reconstruction quality.

Iterative reconstruction methods~\cite{sart_andersen,art,sirt} cast CBCT as an inverse problem by repeatedly updating voxel intensities to minimize projection discrepancies. 
These methods allow the integration of regularizers, such as total variation (TV)~\cite{demircan2017discretized,xu2019total,du2016evaluation}, which improve edge preservation and noise robustness. 
Yet, their reliance on voxel discretization and repeated forward and back projections leads to high computational cost, making them impractical for high-resolution CBCT volumes.

To move beyond voxel-based representations, recent work incorporates continuous neural fields. 
SAX-NeRF~\cite{sax_nerf} uses transformer-based attention along projection rays to reconstruct high-quality volumes, but optimization remains slow and memory-intensive. 
NAF~\cite{naf_zha} and NeAT~\cite{neat} improve efficiency using octrees and the instant-NGP~\cite{instantngp_muller}, achieving faster reconstructions but still exhibiting oversmoothing and blurry structures. 
R$^2$-Gaussian~\cite{r2_gaussian} introduces Gaussian splatting with differentiable projection and Beer–Lambert modeling, providing both speed and fine detail. 
However, the method assumes a normalized volume space and relies on a monochromatic attenuation model, which prevents accurate modeling of real polychromatic X-ray physics and limits its ability to handle metal-induced nonlinearity.

\mparagraph{Metal artifact reduction}
Early metal artifact reduction  techniques~\cite{limar,nmar} identify metal-corrupted regions in the sinogram and replace them with interpolated neighboring values to suppress streaking artifacts. 
Although simple and fast, these heuristic approaches are highly sensitive to metal size, shape, and placement, which leads to inconsistent behavior across scans and an inability to model the underlying physics.

Learning-based post-processing methods~\cite{acdnet,dicdnet,oscnet,adn,indudonet} attempt to remove artifacts directly from reconstructed slices using neural networks. 
While they can improve visual quality within the distribution of the training data, their dependence on supervised examples limits generalization under different imaging geometries, object compositions, or acquisition conditions.

More physically grounded approaches incorporate differentiable CBCT models to enable physics-aware MAR~\cite{sax_nerf,r2_gaussian}. 
Polyner~\cite{polyner} embeds the source spectrum and metal masks as priors to estimate energy-dependent attenuation, but its reliance on metal segmentation and its NeRF backbone results in amplified noise and slow convergence. 
Diner~\cite{diner} simplifies the problem using an energy-independent attenuation parameter, which reduces complexity at the cost of missing key polychromatic behavior. 
Park et al.~\cite{yonsei} assume a linearly energy-dependent model, avoiding metal masks but oversimplifying the attenuation physics, which leaves noticeable residual artifacts.

In contrast, our method introduces a splat-based, physics-driven polychromatic projection and differentiable reconstruction framework that does not require metal masks or computationally heavy volumetric rendering networks. 
Furthermore, our formulation naturally supports incorporating image-domain priors. By leveraging a compact material representation within Gaussian splatting, our approach preserves sharp structural boundaries while accurately modeling metal-induced nonlinear attenuation, resulting in more effective and efficient artifact reduction.

\section{Background}
\label{sec:background}

\mparagraph{Polychromatic forward projection}
X-ray photons attenuate according to the Beer–Lambert law:
\begin{equation}
	\label{eq:beer_lambert}
	I = I_0 \exp\!\left( -\int_L \mu(l)\, dl \right),
\end{equation}
where $I_0$ is the source intensity, $\mu(l)$ is the linear attenuation coefficient along the ray path $L$, and $I$ is the detector measurement. Taking the logarithm yields the standard monochromatic projection model:
\begin{equation}
	\label{eq:proj_mono}
	P = -\log \frac{I}{I_0} = \int_L \mu(l)\, dl.
\end{equation}
Real X-ray sources, however, emit polychromatic spectra, and both the source and attenuation depend on photon energy. The monochromatic model therefore fails to describe the true physical process. The correct formulation integrates attenuation over the entire energy spectrum~$\mathcal{E}$:
\begin{equation}
	\label{eq:proj_poly}
	P = -\log \int_\mathcal{E} \eta(E)\,\exp\!\left( -\int_L \mu(l,E)\, dl \right)\, dE,
\end{equation}
where $\eta(E)$ denotes the spectral distribution and filtration, and $\mu(l,E)$ is the energy-dependent linear attenuation coefficient (LAC). This coupling between energy and material composition produces the characteristic beam-hardening behavior observed in polychromatic CT and creates a strong physical prior for inferring material-specific attenuation.

\mparagraph{Material attenuation}
The linear attenuation coefficient can be factorized into mass density $\rho$ and the mass attenuation coefficient (MAC) $\mu_{\rho}(E)$ \cite{nist}:
\begin{equation}
	\label{eq:lac_decomp}
	\mu(E) = \rho\,\mu_{\rho}(E).
\end{equation}
The MAC is an intrinsic quantity determined solely by elemental composition and photon energy. It does not depend on the material’s geometry or microstructure, and tabulated MAC curves are available for most materials through NIST \cite{nist}. The density $\rho$, in contrast, varies with physical structure, such as variations in bone density across individuals or the difference between soft tissue and cortical tissue. This decomposition separates a fixed, energy-dependent material signature from a variable, scene-specific density term. In our Gaussian-splatting representation, this structure-dependent density is encoded in the density parameter of each Gaussian primitive, enabling us to model heterogeneous attenuation fields.

\mparagraph{Splat-based reconstruction}
In R$^2$-Gaussian~\cite{r2_gaussian}, the attenuation field is modeled as a continuous sum of anisotropic Gaussian primitives:
\begin{equation}
	\label{eq:r2gs_atten}
	\begin{aligned}		
		\mu(\mathbf{x}) = \sum_{i}^{M} \delta_i \exp \left( -\frac{1}{2} (\mathbf{x} - \mathbf{p}_i)^{T} \mathbf{\Sigma}_i^{-1} (\mathbf{x} - \mathbf{p}_i) \right),
	\end{aligned}
\end{equation}
where $\mathbf{x}$ is a spatial query point, $M$ is the number of Gaussians, $\delta_i$ is the density of the $i$th Gaussian, and $\mathbf{p}_i$ and $\mathbf{\Sigma}_i$ denote its center and covariance.

The forward projection of a ray is computed by accumulating the contributions of all Gaussians along the ray path:
\begin{equation}
	\label{eq:r2gs_proj}
	\scalebox{0.85}{$
		\begin{aligned}		
			P \left(\hat{\mathbf{x}}\right) &= \int_L \mu(l)\, dl \\
			&\approx \sum_{i}^M \sqrt{\frac{2\pi |\tilde{\mathbf{\Sigma}}_i|}{|\hat{\mathbf{\Sigma}}_i|}} \, \delta_i \exp\left( -\frac{1}{2} (\hat{\mathbf{x}} - \hat{\mathbf{p}}_i)^T \hat{\mathbf{\Sigma}}_i^{-1} (\hat{\mathbf{x}} - \hat{\mathbf{p}}_i) \right),
		\end{aligned}
		$}
\end{equation}
where $\hat{\mathbf{x}}$ is the pixel location, $\tilde{\mathbf{\Sigma}}_i$ is the view-space covariance, $\hat{\mathbf{\Sigma}}_i$ is the projected covariance, and $\hat{\mathbf{p}}_i$ is the projected center. The parameters $\delta_i$, $\mathbf{p}_i$, and $\mathbf{\Sigma}_i$ are optimized so that Equation~\eqref{eq:r2gs_proj} matches the measured projections. After optimization, the final attenuation field is evaluated using Equation~\eqref{eq:r2gs_atten}, yielding a continuous reconstruction.

\section{Method}
\label{sec:method}
\mparagraph{Overview}

\begin{figure*}[h]
	\centering
		\vspace{-3mm}
	\includegraphics[width=1.0\linewidth]{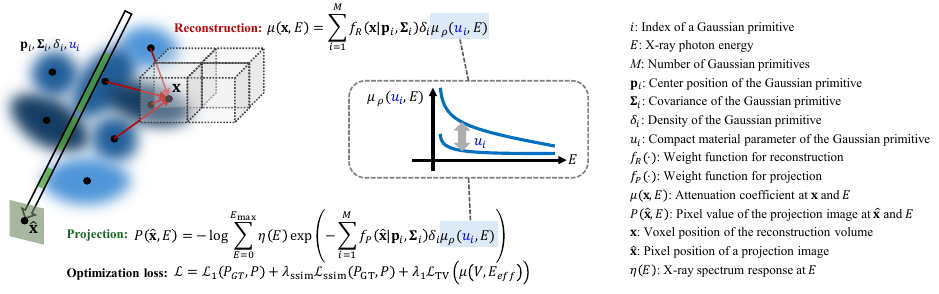}%
	\vspace{-3mm}
	\caption{\label{fig:method_overview}
		Method overview. Instead of using the conventional monochromatic attenuation assumption, we introduce an energy-dependent formulation and additional compact material parameters for each Gaussian primitive to simulate energy-dependent attenuation. 
		These parameters are jointly optimized along with the baseline Gaussian parameters during reconstruction.}
	\vspace{-4mm}
\end{figure*}
Our method extends Gaussian-splatting based CBCT reconstruction~\cite{r2_gaussian} to perform metal artifact reduction by explicitly modeling polychromatic X-ray physics. 
Figure~\ref{fig:method_overview} summarizes the overall pipeline. Instead of assuming a monochromatic source, we adopt the full polychromatic projection model in Equation~\eqref{eq:proj_poly}, which integrates energy-dependent attenuation along each ray. Directly optimizing this formulation, however, is challenging due to its nonlinear coupling between material properties and the X-ray spectrum.

To make this optimization tractable, we assign each Gaussian primitive a compact material parameter that governs its mass attenuation behavior across energy. 
This parameter controls a low-dimensional approximation of the MAC curve, allowing us to express the polychromatic forward model with significantly fewer variables while maintaining physical realism. 
As a result, our approach can jointly optimize geometric parameters and material-dependent attenuation in a mask-free manner, enabling efficient and physically consistent metal artifact reduction within the Gaussian-splatting framework.

\mparagraph{Compact material attenuation}
In medical imaging, metallic implants introduce strong energy-dependent attenuation. Among clinically used metals, titanium and iron are most common, with aluminum, chromium, cobalt, and nickel also appearing in specialized applications~\cite{boas2012ct,hokamp2020quantification,selles2024advances}. Biological tissues, 
in contrast, are largely composed of water, protein, lipid, and carbohydrate~\cite{valentin2002basic,white1987average}, which results in a comparatively narrow range of mass attenuation coefficients.

Empirically, the MAC curves of common biological tissues and clinically relevant metals occupy a low-dimensional manifold. Although the absolute magnitude differs substantially between soft tissue and metal, the shape of the MAC curve across energy remains smooth and highly correlated. 
Leveraging this observation, we approximate the MAC of each Gaussian primitive using a quadratic B\'{e}zier curve whose control points represent the minimum, intermediate, and maximum MAC vectors ($\mathbf{b}_{s}$, $\mathbf{b}_{m}$, $\mathbf{b}_{f}$) observed across these materials:
\begin{equation}
	\label{eq:quadratic_bezier}
	\scalebox{0.9}{$
		\mu_{\rho}(u_i, E)
		= (1 - u_i)^2 \,\mathbf{b}_{s}(E)
		+ 2(1 - u_i)u_i\, \mathbf{b}_{m}(E)
		+ u_i^2 \,\mathbf{b}_{f}(E),
		$}
\end{equation}
where $u_i \in [0,1]$ is a continuous scalar that smoothly interpolates along this material manifold.

This compact representation offers three key benefits. 
First, it captures the dominant variation of real MAC curves across energy, since the difference between materials lies primarily in scale and curvature rather than in complex high-frequency patterns. 
Second, it provides a compact and smoothly varying parameterization that stabilizes the optimization of polychromatic attenuation within the Gaussian framework. 
Third, by encoding MAC variation in one scalar, it avoids spatially inconsistent attenuation estimates that arise when optimizing a high-dimensional material vector at every primitive.

\begin{figure}[tp]
	\centering
		\vspace{-1mm}\includegraphics[width=1.0\linewidth]{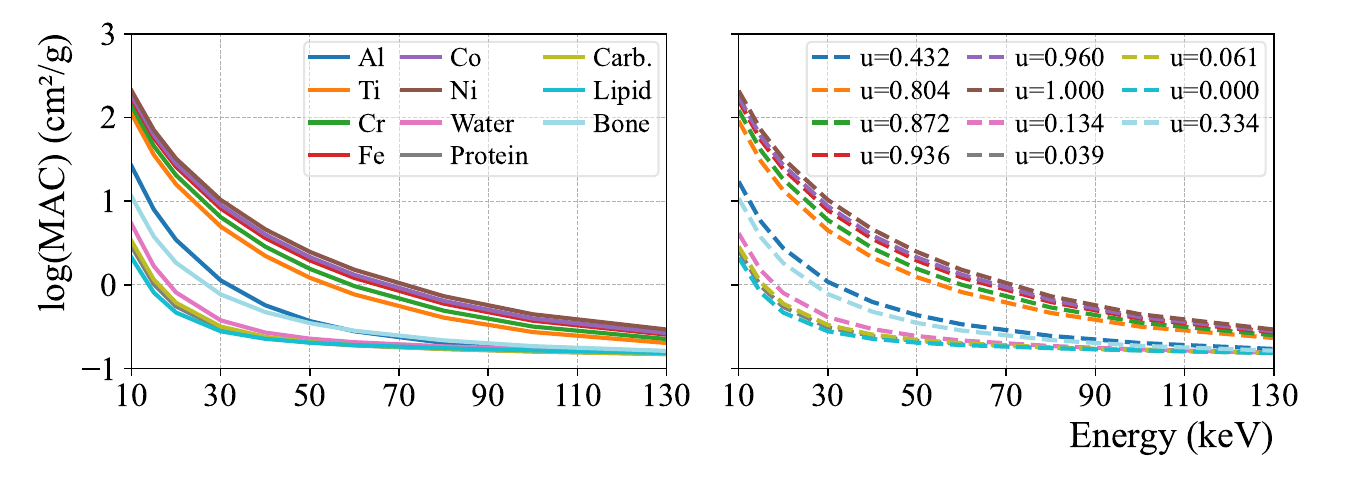}%
	\vspace{-4mm}
	\caption{\label{fig:observation}
		(a) Representative mass attenuation coefficients~(MACs) from the NIST database~\cite{nist} of biological and metallic materials.
		(b) A quadratic B\'{e}zier curve defined by minimum, intermediate, and maximum MAC vectors closely approximates the MACs of real materials in (a), confirming that the real MACs lie on a smooth, low-dimensional manifold.
	}
		\vspace{-4mm}
\end{figure}

To validate this formulation, we compare real MAC curves from NIST with their Bézier-approximated counterparts (Figure~\ref{fig:observation}). 
The interpolated curves reproduce both the magnitude and the shape of the energy-dependent attenuation with high fidelity, indicating that a scalar parameter $u_i$ is sufficient to approximate the physically relevant variation of MACs. 
Prior work has also observed similar low-dimensional structure in energy-dependent attenuation for related tasks~\cite{long2014multi}, supporting the feasibility of our compact modeling approach.

\mparagraph{Polychromatic forward projection}
The baseline approach~\cite{r2_gaussian} reformulates the Beer--Lambert law into a Gaussian-based forward projection model (Equation~\eqref{eq:r2gs_proj}), but it assumes a \emph{monochromatic} X-ray source and therefore cannot capture the energy-dependent attenuation responsible for metal artifacts. 
We address this limitation by extending the formulation to a fully polychromatic model (Equation~\eqref{eq:gs_poly_proj}), where the X-ray spectrum is explicitly integrated over energy:
\begin{equation}
	\label{eq:gs_poly_proj}
	\scalebox{0.8}{$
		\begin{aligned}
			P(\hat{\mathbf{x}}, E)
			= -\log
			\sum_{E=0}^{E_{\max}}
			\eta(E)\,
			\exp\!\left(
			-\sum_{i=1}^{M}
			f_P(\hat{\mathbf{x}} \mid \mathbf{p}_i, \mathbf{\Sigma}_i)\;
			\delta_i\;
			\mu_{\rho}(u_i, E)
			\right),
		\end{aligned}
		$}
\end{equation}
where $\eta(E)$ is the X-ray spectrum response, and 
$f_P(\hat{\mathbf{x}} \mid \mathbf{p}_i, \mathbf{\Sigma}_i)$ is the projection weight of the $i$-th Gaussian:
$
f_P(\hat{\mathbf{x}} \mid \mathbf{p}_i, \mathbf{\Sigma}_i)
= 
\sqrt{\frac{2\pi |\tilde{\mathbf{\Sigma}}_i|}{|\hat{\mathbf{\Sigma}}_i|}} 
\exp\!\left(
-\frac{1}{2}(\hat{\mathbf{x}} - \hat{\mathbf{p}}_i)^{T} 
\hat{\mathbf{\Sigma}}_i^{-1} 
(\hat{\mathbf{x}} - \hat{\mathbf{p}}_i)
\right)
$.
Here, $\delta_i$ denotes the density of the primitive, and $\mu_{\rho}(u_i,E)$ is the energy-dependent attenuation model controlled by its \emph{compact material parameter} $u_i$. 
This formulation embeds a differentiable polychromatic Beer--Lambert law directly into the Gaussian splatting framework.

To model the X-ray spectrum, we use $\eta(E)$ generated by the SPEKTR simulator~\cite{spektr}, following prior polychromatic reconstruction studies~\cite{polyner,diner}. 
By jointly optimizing the compact material parameter $u_i$ along with all Gaussian primitives, our method adaptively captures diverse attenuation behaviors across tissue and metal, which enables effective suppression of metal-induced artifacts.

The attenuation field is reconstructed by aggregating all Gaussian contributions:
\begin{equation}
	\label{eq:gs_poly_atten}
	\mu(\mathbf{x}, E)
	= \sum_{i=1}^{M}
	f_R(\mathbf{x} \mid \mathbf{p}_i, \mathbf{\Sigma}_i)\;
	\delta_i\;
	\mu_{\rho}(u_i, E),
\end{equation}
where the reconstruction weight is
\[
f_R(\mathbf{x} \mid \mathbf{p}_i, \mathbf{\Sigma}_i)
=
\exp\!\left(
-\frac{1}{2}(\mathbf{x} - \mathbf{p}_i)^{T}
\mathbf{\Sigma}_i^{-1}
(\mathbf{x} - \mathbf{p}_i)
\right).
\]
This energy-aware reconstruction enables our method to accurately recover attenuation volumes across both biological tissue and metal regions, providing a physically consistent foundation for metal artifact reduction.
Furthermore, we design the reconstruction pipeline to be fully differentiable, enabling seamless integration of image-domain priors. Detailed derivations of the loss gradients with respect to the intermediate variables in the reconstruction pipeline are provided in the supplementary material.

\mparagraph{Optimization}
We optimize the per-Gaussian parameters~$\{ \mathbf{p}_i, \bm{\Sigma}_i, \delta_i, u_i  \mid i = 1, \dots, M \}$.
where $u_i$ is the compact material parameter that controls the energy-dependent attenuation of each primitive. 
The overall objective combines a pixel-wise $L_1$ loss, a structural similarity (SSIM) loss, and a total variation (TV) loss computed over randomly sampled voxel locations, following the baseline splat-based reconstruction~\cite{r2_gaussian}:
\begin{equation}
	\label{eq:ours_loss}
	\scalebox{0.9}{$
	\begin{aligned}		
		\mathcal{L}_{\text{total}} = \mathcal{L}_1(P_{\text{GT}}, P) + \lambda_{0} \mathcal{L}_{\text{SSIM}}(P_{\text{GT}}, P) + \lambda_{1} \mathcal{L}_{\text{TV}}(\mu(V,E_{eff})),
	\end{aligned}
	$}
\end{equation}
Here, $V$ denotes voxel-grid query points. We set $\lambda_{0}=0.25$ and $\lambda_{1}=3.0$ in all experiments. Full gradient expressions for $P$ and $V$ with respect to the Gaussian parameters are provided in the supplemental material, enabling end-to-end backpropagation through our CUDA-based differentiable projection pipeline.

After optimization, the final voxelized attenuation field is obtained using Equation~\eqref{eq:gs_poly_atten}. The reconstructed image is then computed at the effective energy
$E_{\text{eff}} = \sum_{i=1}^{N} \eta(E_i) E_i$,
where $N$ is the number of spectral bins, following prior work~\cite{diner}.

\section{Evaluation}
\label{sec:results}
\mparagraph{Implementation}
We implement our method in PyTorch~\cite{pytorch} with CUDA~\cite{cuda}. The B\'{e}zier MAC basis is constructed from the NIST~\cite{nist} attenuation curves of water, iron, and aluminum, assuming an operating spectrum of 10–90~keV. We use $N=15$ uniformly spaced energy samples for spectrum integration and adopt the SPEKTR-generated spectral response $\eta(E)$ following Polyner~\cite{polyner}.
Gaussian initialization follows R$^2$-Gaussian~\cite{r2_gaussian}. 

\begin{figure*}[!t]
	\centering
\vspace{-5mm}
	\includegraphics[width=0.85\linewidth]{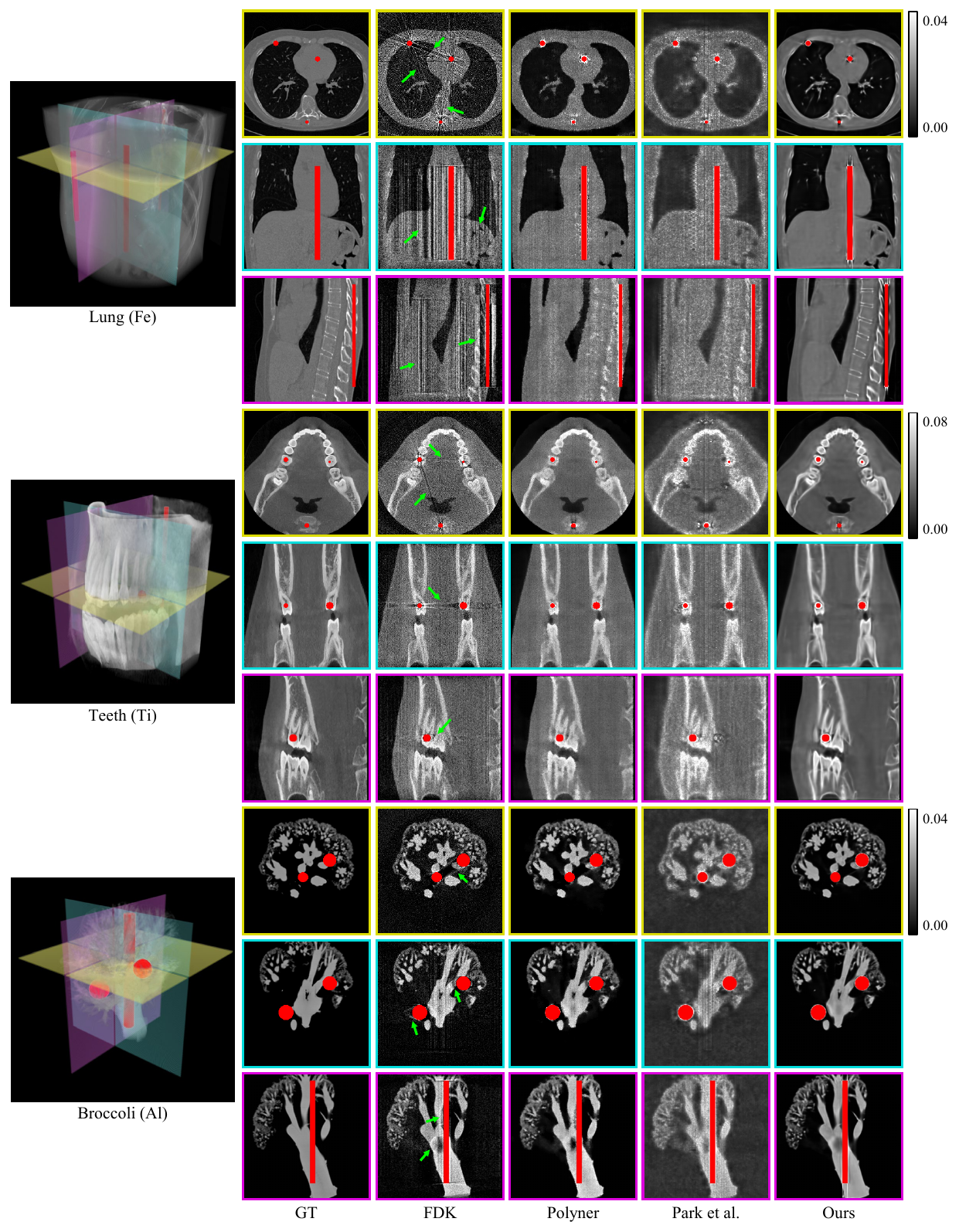}%
	\vspace{-4mm}
	\caption{\label{fig:results_synthetic_comparison}
Axial (yellow), coronal (pink), and sagittal (green) slices from the three synthetic scenes are shown, with metal regions highlighted in red. FDK produces severe streaking and shading artifacts. Polyner and Park et al. reduce some streaks but suffer from detail loss, contrast distortion, and heavy blurring (Park et al. displayed with a 0.12 intensity cap). In contrast, our method preserves fine structural details and effectively eliminates metal-induced artifacts, yielding the most faithful reconstruction across all scenes.}
\vspace{-5mm}
\end{figure*}

\subsection{Validation}
\mparagraph{Dataset}
We evaluate our method on three synthetic CBCT phantoms---\textit{Lung}, \textit{Teeth}, and \textit{Broccoli}---constructed from the LIDC~\cite{lidc}, X-plant~\cite{xplant}, and ZCB100~\cite{zcb100} datasets, each containing inserted metal objects. The datasets are generated using the synthetic projection pipeline introduced in prior work~\cite{eg2026_mar}.

\mparagraph{Evaluation on synthetic dataset}
Figure~\ref{fig:results_synthetic_comparison} shows qualitative comparisons on three synthetic scenes. FDK~\cite{feldkamp1984practical} produces strong streak artifacts due to its monochromatic assumption, and Polyner~\cite{polyner} suppresses streaks but loses fine structures. Park et al.~\cite{yonsei} yields heavily blurred results with incorrect intensity levels.
Our method preserves structural details—clearly separating bone and soft tissue in \textit{Lung} and retaining high-frequency textures in \textit{Broccoli}—while effectively removing metal artifacts.
As reported in Table~\ref{tb:synthetic_experiment}, our approach achieves the highest 3D PSNR and SSIM across all phantoms. LIMAR~\cite{limar}, NMAR~\cite{nmar}, and Polyner improve over FDK but remain limited, whereas supervised baselines~\cite{acdnet,dicdnet,oscnet} fail to generalize due to domain gaps.
We provide additional results on two synthetic scenes—a pancreas with titanium (Ti) and a pepper with iron (Fe)—in the supplementary material.

\subsection{Real Experiment}
\mparagraph{Dataset}
We acquire real CBCT scans using a Bruker SKYSCAN 1273 system at 90 kVp with 1.0 mm aluminum filtration.
We construct four real datasets using readily available objects that mimic key anatomical characteristics: garlic (complex structures), enoki mushroom and blueberry (fine textures), and avocado (multiple internal materials). To represent soft-tissue and bone characteristics, we additionally employ the chicken dataset from existing work~\cite{eg2026_mar}. 
Furthermore, to evaluate our method on a human head–like structure, we also include the walnut dataset from the same work.
Each specimen is scanned both with and without metal to capture metal-induced beam hardening, with the metal-free scans serving as a reference.
Additional system details are provided in the supplemental material.

\begin{figure*}[!t]
	\centering
		\vspace{-2mm}
		\includegraphics[width=0.87\linewidth]{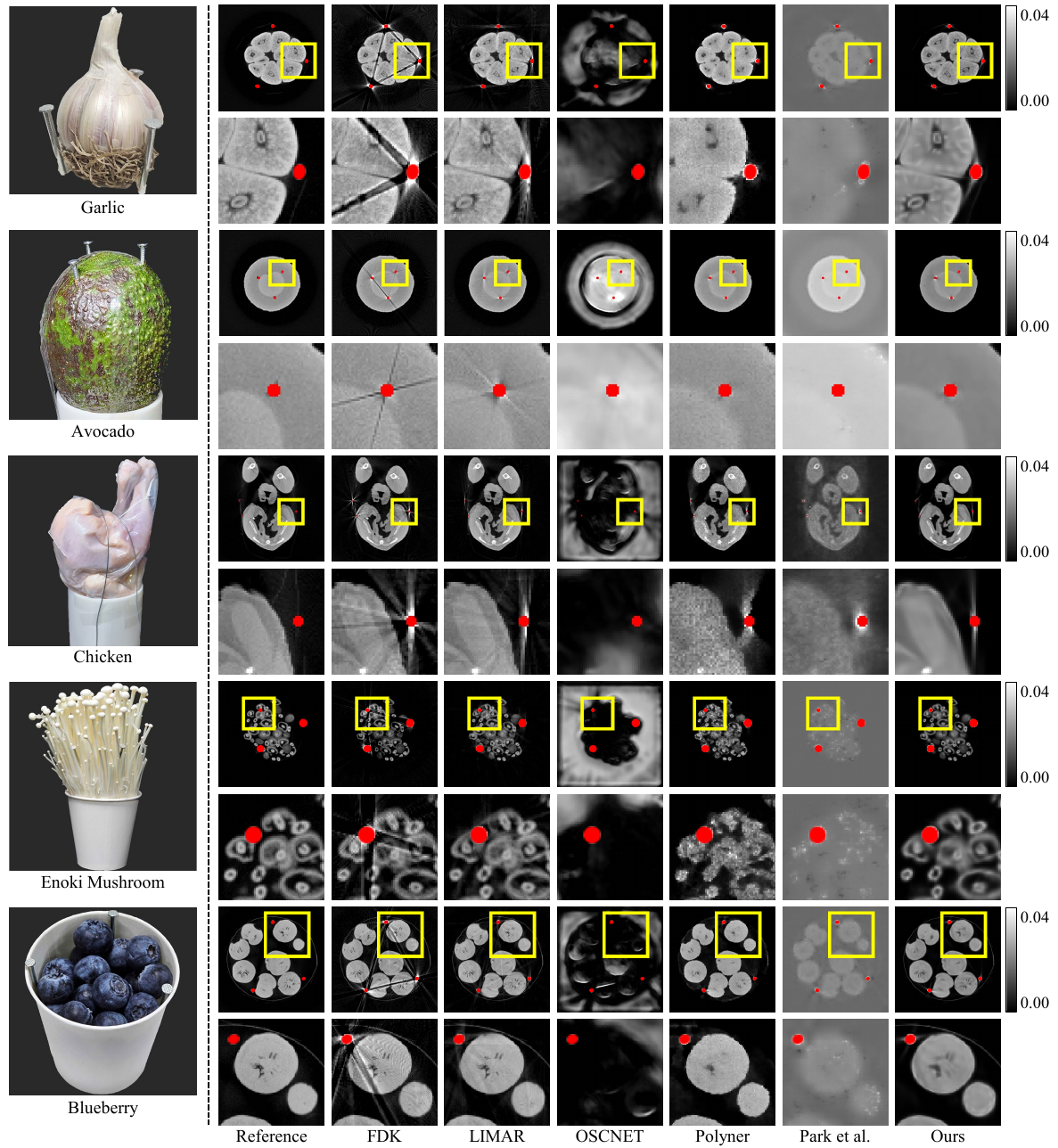}
	\vspace{-2mm}
	\caption{\label{fig:results_real_comparison}
Real-data reconstruction results.
FDK~\cite{feldkamp1984practical} produces strong streaking and shading artifacts. LIMAR~\cite{limar} reduces some of these effects but leaves noticeable residuals. OSCNet~\cite{oscnet} fails to generalize to unseen acquisition conditions and produces incorrect intensity distributions. Polyner~\cite{polyner} suppresses streaks but oversmooths fine structural details. Park et al.~\cite{yonsei} yields heavily blurred reconstructions and does not recover the correct intensity scale due to its simplified polychromatic model. In contrast, our method effectively removes metal-induced artifacts while preserving material-dependent density variations and high-frequency structural details across all specimens. (For visualization, Park et al. is displayed with an intensity cap of 0.1 for avocado and 0.2 for the remaining scenes.)}
\vspace{-5mm}
\end{figure*}

\mparagraph{Evaluation on real dataset}
Figure~\ref{fig:results_real_comparison} shows reconstruction results on the real-world datasets. Similar to the synthetic experiments, FDK~\cite{feldkamp1984practical} produces strong streaking artifacts due to its monochromatic assumption. LIMAR~\cite{limar} reduces some of the severe artifacts, particularly in the avocado scene, but leaves noticeable residual streaks. OSCNet~\cite{oscnet} struggles to generalize beyond its training distribution, leading to poor overall reconstruction quality.
Polyner~\cite{polyner} achieves strong artifact suppression, yet it oversmooths important structural details, failing to capture the garlic’s fine internal texture and the chicken’s soft-tissue variations. Park et al.~\cite{yonsei} also produce overly blurred reconstructions, losing high-frequency content across all scenes.
In contrast, our method consistently removes metal-induced artifacts while preserving material-dependent density variations and high-frequency structural details, yielding the most faithful reconstructions across all real-world cases.

\subsection{Computation Time}
We compare the computation time of our method with state-of-the-art joint reconstruction and metal artifact reduction approaches in Table~\ref{tb:ablation_computation_time}. All measurements are performed on an Intel Xeon 4214R CPU and an NVIDIA RTX A6000 GPU. Across all scenes, our method achieves an order-of-magnitude speedup, demonstrating the computational advantage of the splat-based formulation and the compact material parameterization used in our polychromatic model.

\begin{table}[tp]
	\centering	\caption{\label{tb:ablation_computation_time}
{Efficiency comparison for joint reconstruction and metal artifact reduction methods. Ours is consistently faster across both synthetic and real datasets, achieving significant speedups over existing polychromatic and neural-field–based methods.}
	}
	\vspace{-2mm}
	
	\resizebox{1.0\linewidth}{!}{
		\begin{tabular}{l l r r r}
			\toprule
			Dataset & Scene 
			& Polyner~\cite{polyner} 
			& Park et al.~\cite{yonsei} 
			& Ours \\
			\midrule
			
			\multirow{3}{*}{Synthetic}
			& Broccoli & 1h 50m 02s & 1h 48m 49s & \textbf{28m 57s} \\
			& Lung     & 1h 14m 34s & 1h 12m 38s & \textbf{41m 40s} \\
			& Teeth    & 1h 15m 44s & 1h 13m 41s & \textbf{41m 14s} \\
			\midrule
			
			\multirow{5}{*}{Real}
			& Walnut       & 1h 11m 58s & 1h 10m 20s & \textbf{30m 10s} \\
			& Garlic       & 1h 30m 19s & 1h 26m 34s & \textbf{19m 54s} \\
			& Avocado      & 1h 56m 49s & 1h 50m 24s & \textbf{27m 02s} \\
			& Chicken      & 2h 02m 55s & 2h 09m 43s & \textbf{23m 36s} \\
			& Enoki Mushroom     & 1h 33m 10s & 1h 27m 54s & \textbf{21m 11s} \\
			& Blueberry  & 1h 33m 12s & 1h 28m 12s & \textbf{18m 52s} \\
			\bottomrule
		\end{tabular}
	}
	\vspace{-3mm}
\end{table}

\begin{table}[pt]
	\centering
	\caption{
		\label{tb:synthetic_experiment}
Quantitative results on the synthetic CBCT datasets.
We report 3D PSNR and SSIM for the Lung, Teeth, and Broccoli scenes (Figure~\ref{fig:results_synthetic_comparison}). Our method achieves the highest reconstruction accuracy across all cases, demonstrating strong metal-artifact suppression and superior preservation of structural detail.	}
	\vspace{-3mm}
	\resizebox{1.0\linewidth}{!}{
		\begin{tabular}{l c c c c c c}
			\toprule
			\multirow{2}{*}{Method} 
			& \multicolumn{2}{c}{Lung} 
			& \multicolumn{2}{c}{Teeth} 
			& \multicolumn{2}{c}{Broccoli} \\
			\cmidrule(lr){2-3} \cmidrule(lr){4-5} \cmidrule(lr){6-7}
			& PSNR3D & SSIM3D & PSNR3D & SSIM3D & PSNR3D & SSIM3D \\
			\midrule
			FDK~\cite{feldkamp1984practical} & 17.21 & 0.905 & 17.71 & 0.885 & 18.00 & 0.963 \\
			LIMAR~\cite{limar}               & 20.02 & 0.947 & 18.47 & 0.899 & 18.65 & 0.965 \\
			NMAR~\cite{nmar}                 & 20.22 & 0.950 & 18.58 & 0.900 & 20.40 & 0.977 \\
			ACDNet~\cite{acdnet}             & 13.89 & 0.655 & 9.21  & 0.436 & 14.36 & 0.939 \\
			DICDNet~\cite{dicdnet}           & 13.48 & 0.622 & 8.73  & 0.401 & 13.99 & 0.929 \\
			OSCNet~\cite{oscnet}             & 13.39 & 0.615 & 8.67  & 0.392 & 13.81 & 0.922 \\
			Polyner~\cite{polyner}           & 20.63 & 0.977 & 20.74 & 0.970 & 21.60 & 0.990 \\
			Park et al.~\cite{yonsei}        & 18.80 & 0.887 & 14.90 & 0.768 & 0.07  & 0.201 \\
			{Ours} & \textbf{28.96} & \textbf{0.994} 
			& \textbf{27.40} & \textbf{0.993} 
			& \textbf{27.76} & \textbf{0.997} \\
			\bottomrule
		\end{tabular}
	}
	\vspace{2mm}
	\centering	
	\caption{\label{tb:ablation_spectrum_count}
		Effect of the number of spectral components on reconstruction performance.}
	\vspace{-3mm}
	\resizebox{0.9\linewidth}{!}{%
		\begin{tabular}{l|cccc}
			\hline
			\# of Spec. Comp. ($N$) & 7 & 15 & 31 & 63 \\
			\hline
			PSNR3D & 28.01 & \textbf{28.04} & 27.93 & 27.99 \\
			SSIM3D & 0.994 & \textbf{0.995} & 0.994 & 0.994 \\
			\hline
		\end{tabular}
	}%
	\vspace{-4mm}
\end{table}

\section{Ablation}

\mparagraph{Component Contribution}
We evaluate the impact of our polychromatic model and TV regularization. As shown in Table~\ref{tb:ablation_component}, the polychromatic model already yields a large improvement over the baseline~\cite{r2_gaussian}, and adding TV provides a small additional gain.
\begin{table}[h]
	\centering
	\caption{Ablation of the polychromatic model and TV regularization.
Both components improve reconstruction quality over the baseline~\cite{r2_gaussian}.}
\vspace{-3mm}
	\label{tb:ablation_component}
	\resizebox{0.8\linewidth}{!}{%
		\begin{tabular}{l|cc}
			\hline
			Module & PSNR3D & SSIM3D \\ 
			\hline
			Baseline & 23.22 & 0.984 \\
			Baseline + Poly (Ours) & 27.97 & 0.994 \\
			Baseline + Poly + TV (Ours) & \textbf{28.04} & \textbf{0.995} \\
			\hline
		\end{tabular}}%
	\vspace{-5mm}
\end{table}

\mparagraph{Number of spectrum components}
We evaluate the effect of the number of spectral components 
$N$ used for polychromatic integration. As shown in Table~\ref{tb:ablation_spectrum_count}, reconstruction quality remains stable across a wide range of $N$, 
indicating that our compact material model is robust to the spectral sampling density. 
The differences between 7, 15, 31, and 63 samples are small, which suggests that the underlying attenuation behavior is well captured even with relatively coarse sampling. 
We choose $N=15$ as a practical operating point, since it provides the best balance between numerical accuracy and computational efficiency, while further increasing $N$ yields negligible gains and slightly higher runtime.

\section{Limitation}

Our compact material model captures the dominant variation in energy-dependent attenuation, yet it cannot represent highly unusual materials or compounds whose MAC curves deviate significantly from the low-dimensional manifold shown in Figure~\ref{fig:observation}. 
In practice, the X-ray spectrum cannot be measured directly because energy-integrating detectors are used. 
Following common practice and prior work~\cite{polyner}, we therefore rely on physics-based spectrum simulation using SPEKTR~\cite{spektr}. 
Nevertheless, inaccuracies in spectrum or filtration modeling may introduce residual artifacts or global attenuation bias. 
In addition, while Gaussian splatting provides substantial efficiency, its performance depends on the density and placement of primitives, which may require tuning for extremely large volumes or highly anisotropic structures. 
Finally, our evaluation focuses on static CBCT scans; extending the method to dynamic or limited-angle acquisitions remains an open direction for future work.

\section{Conclusion}
\label{sec:conclusion}
We introduced a CBCT reconstruction framework that mitigates metal-induced beam-hardening artifacts through explicit polychromatic modeling and a Gaussian-splatting formulation.
Our method jointly estimates both scene structure and energy-dependent attenuation parameters, all without relying on metal masks or paired supervision.
Comprehensive experiments on synthetic and real scans demonstrate that our approach surpasses state-of-the-art baselines in both reconstruction fidelity and artifact reduction.
To further facilitate research on artifact-robust reconstruction, we also provide new CBCT datasets featuring realistic metal-induced artifacts.

\appendix
\vspace{-0.1cm}
\section*{Acknowledgements}
\vspace{-0.15cm}
\noindent Min H.~Kim acknowledges the Samsung Research Funding \& Incubation Center (SRFC-IT2402-02), the Korea NRF grant (RS-2024-00357548), the MSIT/IITP of Korea (RS-2022-00155620, RS-2024-00398830, RS-2024-00436680, and 2017-0-00072), the MSIT Advanced GPU Utilization Support Program, and Microsoft Research Asia, with additional support by the KAIST Jang Young Sil Fellow Program.

{
    \small
    \bibliographystyle{ieeenat_fullname}
    \bibliography{bibliography}
}

\clearpage 
\pagestyle{empty} 


\section*{Supplementary material}

# Splat-Based Metal Artifact Reduction in Cone-Beam CT via Compact Attenuation Modeling

Supplementary Material

## 1. Derivation of Backward Propagation

We implement a custom CUDA [1] pipeline that enables automatic differentiation for both the discrete polychromatic forward projection and reconstruction models, both of which are formulated using Gaussian primitives.

### 1.1. Projection Model

Our polychromatic forward projection is expressed as Equation (1):

$$P(\hat{\mathbf{x}}) = -\log \sum_{k=1}^{N} \eta_k \exp\left( -\sum_{i=1}^{M} s_i G_i^P(\hat{\mathbf{x}})\, \delta_i\, \mu_\rho(u_i, k) \right) \tag{1}$$

where

$$s_i = \sqrt{\frac{2\pi|\tilde{\Sigma}_i|}{|\Sigma_i|}},$$

$$G_i^P(\hat{\mathbf{x}}) = \exp\left( -\frac{1}{2}(\hat{\mathbf{x}} - \hat{\mathbf{p}}_i)^T \tilde{\Sigma}_i^{-1} (\hat{\mathbf{x}} - \hat{\mathbf{p}}_i) \right),$$

$$\mu_\rho(u_i, k) = (1-u_i)^2\, \mathbf{b}_\mathrm{s}(k) + 2(1-u_i)u_i\, \mathbf{b}_\mathrm{m}(k) + u_i^2\, \mathbf{b}_\mathrm{f}(k),$$

$$u_i \in [0,1],$$

$$\sum_{k=1}^{N} \eta_k = 1.$$

The baseline method [10] comprises a `Rasterizer_preprocess`, which transforms each Gaussian's position and shape according to the detector pixel coordinates, and a `Rasterizer_render`, which computes the intensity projected onto each detector pixel. While we reuse the original implementation for the `Rasterizer_preprocess`, we design a new `Rasterizer_render` module and manually derive gradients with respect to each intermediate parameter, as shown in Equations (2)–(5):

$$\left.\frac{d\mathcal{L}}{d\delta_i}\right|_{\hat{\mathbf{x}}} = \frac{d\mathcal{L}}{dP} \cdot \frac{dP}{d\delta_i} = \frac{d\mathcal{L}}{dP} \cdot \left(\frac{C}{A}\right) (s_i G_i^P), \tag{2}$$

$$\left.\frac{d\mathcal{L}}{ds_i}\right|_{\hat{\mathbf{x}}} = \frac{d\mathcal{L}}{dP} \cdot \frac{dP}{ds_i} = \frac{d\mathcal{L}}{dP} \cdot \left(\frac{C}{A}\right) (\delta_i G_i^P), \tag{3}$$

$$\left.\frac{d\mathcal{L}}{dG_i^P}\right|_{\hat{\mathbf{x}}} = \frac{d\mathcal{L}}{dP} \cdot \frac{dP}{dG_i^P} = \frac{d\mathcal{L}}{dP} \cdot \left(\frac{C}{A}\right) (s_i \delta_i), \tag{4}$$

$$\begin{aligned}\left.\frac{d\mathcal{L}}{du_i}\right|_{\hat{\mathbf{x}}} &= \frac{d\mathcal{L}}{dP} \left( \sum_{k=1}^{N} \frac{dP}{d\mu_\rho(u_i,k)} \cdot \frac{d\mu_\rho(u_i,k)}{du_i} \right) \\ &= \frac{d\mathcal{L}}{dP} \cdot \left( \frac{s_i G_i^P \delta_i}{A} \right) \sum_{k=1}^{N} B_k \cdot \frac{d\mu_\rho(u_i,k)}{du_i},\end{aligned} \tag{5}$$

where

$$A = \sum_{k=1}^{N} B_k,$$

$$B_k = \eta_k \exp\left( -\sum_{i=1}^{M} s_i\, G_i^P(\hat{\mathbf{x}})\, \delta_i\, \mu_\rho(u_i,k) \right),$$

$$C = \sum_{k=1}^{N} \left( B_k \mu_\rho(u_i,k) \right),$$

$$\begin{aligned}\frac{d\mu_\rho(u_i,k)}{du_i} &= 2\Big[ (1-u_i)(\mathbf{b}_\mathrm{m}(k) - \mathbf{b}_\mathrm{s}(k)) \\ &\quad + u_i(\mathbf{b}_\mathrm{f}(k) - \mathbf{b}_\mathrm{m}(k)) \Big].\end{aligned}$$

All these equations are then implemented in the CUDA backward function.

### 1.2. Reconstruction Model

Our reconstruction is expressed as Equation (6):

$$\mu(\mathbf{x}, k) = \sum_{i=1}^{M} G_i^R(\mathbf{x})\, \delta_i\, \mu_\rho(u_i, k), \tag{6}$$

where

$$G_i^R(\mathbf{x}) = \exp\left( -\frac{1}{2}(\mathbf{x} - \mathbf{p}_i)^T \Sigma_i^{-1} (\mathbf{x} - \mathbf{p}_i) \right).$$

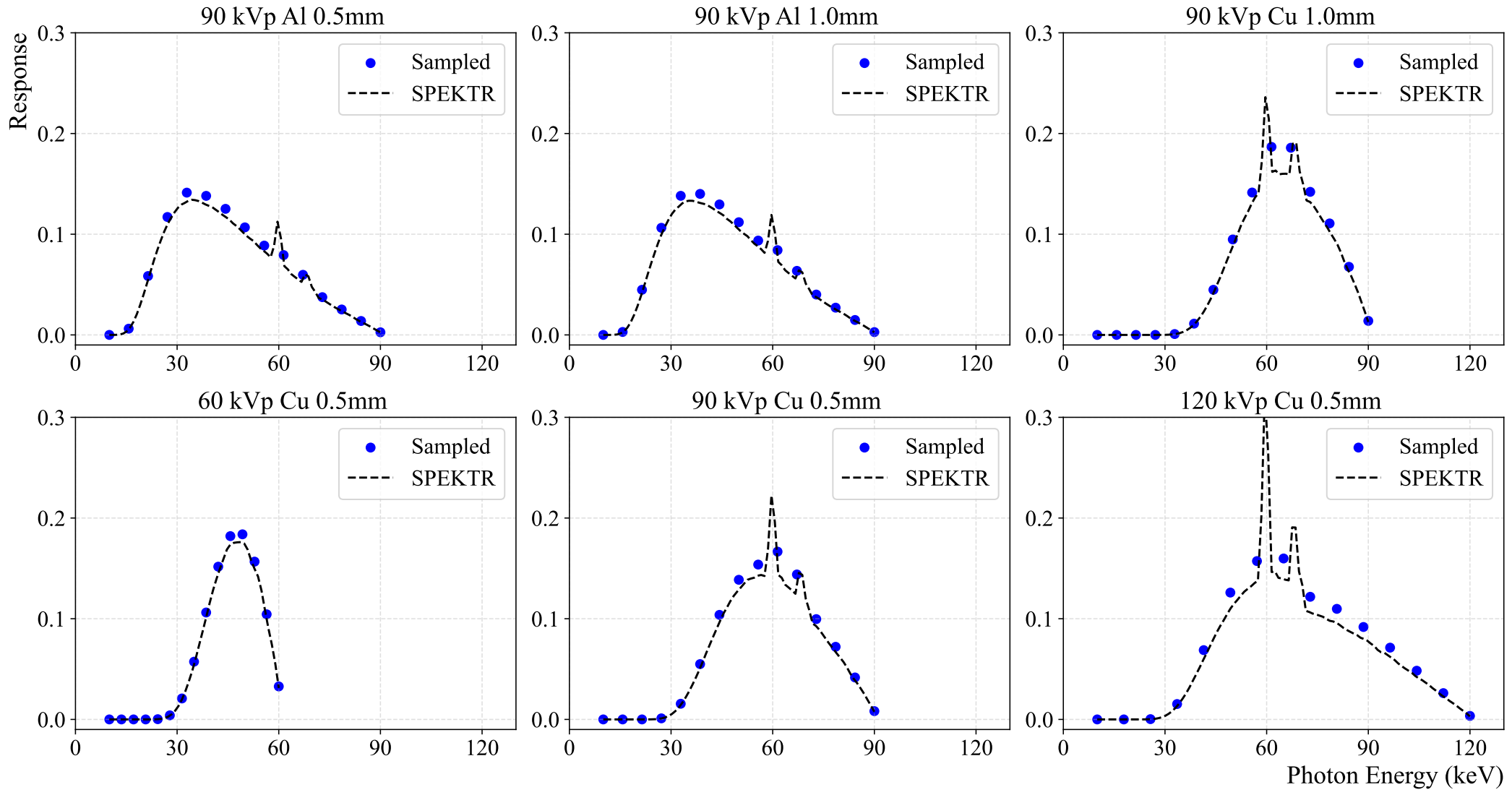


Figure 1. Energy spectra generated using the SPEKTR simulator for various CT system configurations. Each curve represents the normalized photon distribution corresponding to a specific combination of tube voltage and filtration settings. The blue dots denote the 15 uniformly sampled energy components used by our method.

The baseline method [10] comprises a `Voxelizer_preprocess`, which transforms each Gaussian's position and shape according to the voxel grid coordinates, and a `Voxelizer_render`, which computes the intensity accumulated within each voxel. While we reuse the original implementation for the `Voxelizer_preprocess`, we design a new `Voxelizer_render` module and manually derive gradients with respect to each intermediate parameter, as shown in Equations (7)–(9):

$$\left.\frac{d\mathcal{L}}{d\delta_i}\right|_{\mathbf{x}} = \sum_{k=1}^{N} \frac{d\mathcal{L}}{d\mu(k)} \cdot \frac{d\mu(k)}{d\delta_i} = \sum_{k=1}^{N} \frac{d\mathcal{L}}{d\mu(k)} \cdot G_i^R \, \mu_\rho(u_i, k), \tag{7}$$

$$\left.\frac{d\mathcal{L}}{dG_i^R}\right|_{\mathbf{x}} = \sum_{k=1}^{N} \frac{d\mathcal{L}}{d\mu(k)} \cdot \frac{d\mu(k)}{dG_i^R} = \sum_{k=1}^{N} \frac{d\mathcal{L}}{d\mu(k)} \cdot \delta_i \, \mu_\rho(u_i, k), \tag{8}$$

$$\left.\frac{d\mathcal{L}}{du_i}\right|_{\mathbf{x}} = \sum_{k=1}^{N} \frac{d\mathcal{L}}{d\mu(k)} \cdot \frac{d\mu(k)}{d\mu_\rho(u_i,k)} \cdot \frac{d\mu_\rho(u_i,k)}{du_i} = \sum_{k=1}^{N} \frac{d\mathcal{L}}{d\mu(k)} \cdot G_i^R \delta_i \cdot \frac{d\mu_\rho(u_i,k)}{du_i}. \tag{9}$$

All these equations are then implemented in the CUDA backward function.

## 2. X-ray Spectrum

Our method relies on the X-ray spectrum response $\eta(E)$, which depends on the CT system configuration used during data acquisition—specifically, the tube voltage, filter material, and filter thickness. Following prior work [9], we employ the SPEKTR simulator [5] to generate polychromatic X-ray spectra at 1 keV resolution. The simulated spectra are subsequently resampled into 15 uniformly spaced energy components, matching the number of spectral bins used in our model. These interpolated spectral samples are then incorporated into our polychromatic forward projection formulation. Figure 1 visualizes the energy-dependent spectra for the CT settings considered in our experiments.

## 3. Scanning and Reconstruction Configuration

Tables 1–2 summarize the scanning and reconstruction configurations used for both the synthetic and real datasets. To assess the robustness and generality of our method, we evaluate performance across a wide range of acquisition settings, including variations in tube voltage, filtration, geometry, and object scale. These configurations closely follow the parameters of practical CBCT systems and ensure diverse conditions for validating our polychromatic forward-projection and reconstruction framework.

## 4. Fairness of Comparison

Polyner [9] and Park et al. [4] are joint reconstruction and metal artifact reduction methods originally developed for **2D fan-beam** geometry. To enable a fair comparison with our **3D cone-beam** method, we extend both im-

Table 1. Scanning and reconstruction parameters for the synthetic datasets. The left three columns correspond to the main evaluation scenes, and the four rightmost columns denote the additional settings used in the supplementary experiments, including variations in tube voltage and filtration. All other geometric parameters are kept consistent across the datasets.

| **Parameters** | **Synthetic Dataset** | | | | | | |
|---|---|---|---|---|---|---|---|
| | Lung | Teeth | Broccoli | Lung (supple) | Teeth (supple) | Pancreas (supple) | Pepper (supple) |
| Source voltage (kV) | 90 | ← | ← | 60/90/120 | ← | 90 | ← |
| Filter (mm) | Cu(1.0) | Al(0.5) | ← | Cu(0.5) | ← | Cu(1.0) | ← |
| Metal material | Fe | Ti | Al | Fe | Ti | ← | Fe |
| Volume resolution | 256×256×256 | ← | ← | ← | ← | ← | ← |
| Volume size (mm) | 50.0×50.0×50.0 | ← | ← | ← | ← | ← | ← |
| Angle range (°) | [0, 360) | ← | ← | ← | ← | ← | ← |
| Number of angles | 720 | ← | ← | ← | ← | ← | ← |
| Detector resolution | 512×512 | ← | ← | ← | ← | ← | ← |
| Detector size (mm) | 145.485 | ← | ← | ← | ← | ← | ← |
| Source-to-object distance (mm) | 200.962 | ← | ← | ← | ← | ← | ← |
| Source-to-detector distance (mm) | 501.309 | ← | ← | ← | ← | ← | ← |

Table 2. Scanning and reconstruction parameters for the real datasets. Each object is scanned using high-resolution CBCT settings, with variations in object scale, and detector geometry inherent to the acquisition system. These diverse acquisition conditions enable a comprehensive evaluation of metal artifact behavior in real-world scenarios.

| **Parameters** | **Real Dataset** | | | | | |
|---|---|---|---|---|---|---|
| | Walnut | Avocado | Chicken | Garlic | Enoki Mushroom | Blueberry |
| Source voltage (kV) | 90 | ← | ← | ← | ← | ← |
| Source current ($\mu$A) | 166 | ← | 300 | ← | ← | ← |
| Filter (mm) | Al(1.0) | ← | ← | ← | ← | ← |
| Metal material | Fe | ← | ← | ← | ← | ← |
| Volume resolution | 256×256×256 | ← | ← | ← | ← | ← |
| Volume size (mm) | 40.0×40.0×40.0 | 60.0×60.0×60.0 | 120.0×120.0×120.0 | 80.0×80.0×80.0 | ← | ← |
| Angle range (°) | [0, 360) | ← | ← | ← | ← | ← |
| Number of angles | 720 | ← | ← | ← | ← | ← |
| Detector resolution | 512×512 | ← | ← | ← | ← | ← |
| Detector size (mm) | 104.773 | 145.485 | 229.902 | 104.773 | 145.485 | ← |
| Source-to-object distance (mm) | 200.962 | ← | 334.942 | ← | ← | ← |
| Source-to-detector distance (mm) | 501.309 | ← | ← | ← | ← | ← |

Table 3. Quantitative comparison on additional synthetic scenes

| Method | Pancreas (Ti) | | Pepper (Fe) | |
|---|---|---|---|---|
| | PSNR3D | SSIM3D | PSNR3D | SSIM3D |
| FDK | 23.99 | 0.974 | 11.92 | 0.894 |
| LIMAR | 26.76 | 0.985 | 16.33 | 0.952 |
| NMAR | 23.44 | 0.972 | 16.65 | 0.962 |
| ACDNet | 15.41 | 0.776 | 10.77 | 0.879 |
| DICDNet | 15.29 | 0.770 | 10.75 | 0.878 |
| OSCNet | 15.20 | 0.766 | 10.63 | 0.875 |
| Polyner | 28.17 | 0.992 | 22.30 | 0.995 |
| Park et al. | 15.77 | 0.695 | 12.69 | 0.863 |
| Ours | **33.73** | **0.998** | **25.13** | **0.997** |

plementations to the cone-beam setting. LIMAR [2] and NMAR [3] are inherently applicable to 3D reconstruction and are therefore directly comparable. For learning-based methods (ACDNet [7], DICDNet [6], and OSCNet [8]), we use publicly available pre-trained models and apply them slice-wise to FDK-reconstructed volumes, following their original design.

# 5. Robustness to Tube Voltage Variation

To evaluate the stability of our method under different X-ray source energy conditions, we generated synthetic projection images at 60, 90, and 120 kVp and assessed the corresponding reconstruction quality. The Lung and Teeth scenes were simulated using a 0.5 mm Cu filter to match realistic CBCT acquisition settings. As shown in Figure 2, our method maintains consistent reconstruction performance across different tube voltages, demonstrating strong robustness to variations in the underlying X-ray spectrum.

# 6. Additional Discussion

We compute 3D PSNR and SSIM to evaluate reconstruction quality on the synthetic dataset. For PSNR, the ground truth (GT) volume is defined using the effective energy described in the main paper, and—consistent with prior metal artifact reduction studies [4, 9]—the metric is computed over all regions excluding the metal mask. For real datasets, no reliable GT is available; therefore, we report only qualitative comparisons rather than quantitative scores.

# 7. Additional Synthetic Results

We conducted additional experiments on two newly introduced synthetic scenes: a pancreas scene with titanium (Ti)

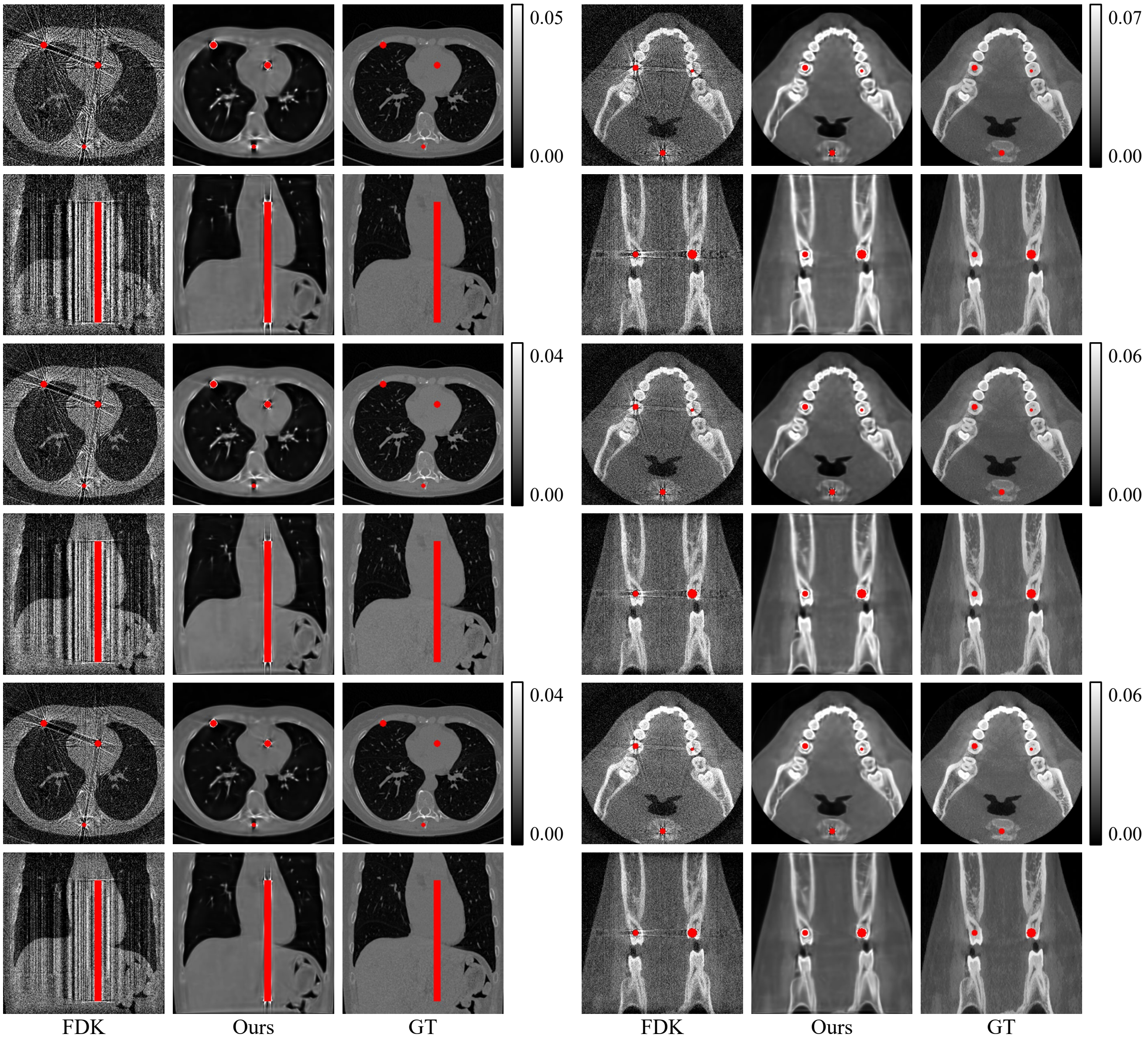


Figure 2. Evaluation of reconstruction robustness under varying tube voltages (60, 90, and 120 kVp). Despite substantial changes in the underlying X-ray spectrum, our method consistently produces high-quality reconstructions. Rows 1–2 show results at 60 kVp, rows 3–4 at 90 kVp, and rows 5–6 at 120 kVp.

and a pepper scene with iron (Fe). Quantitative results, summarized in Table 3, show consistent improvements over all baseline methods. Qualitative results are shown in Figure 4.

## 8. Additional Real Results

We provide additional orthogonal slices for the real-data results in Figure 3.

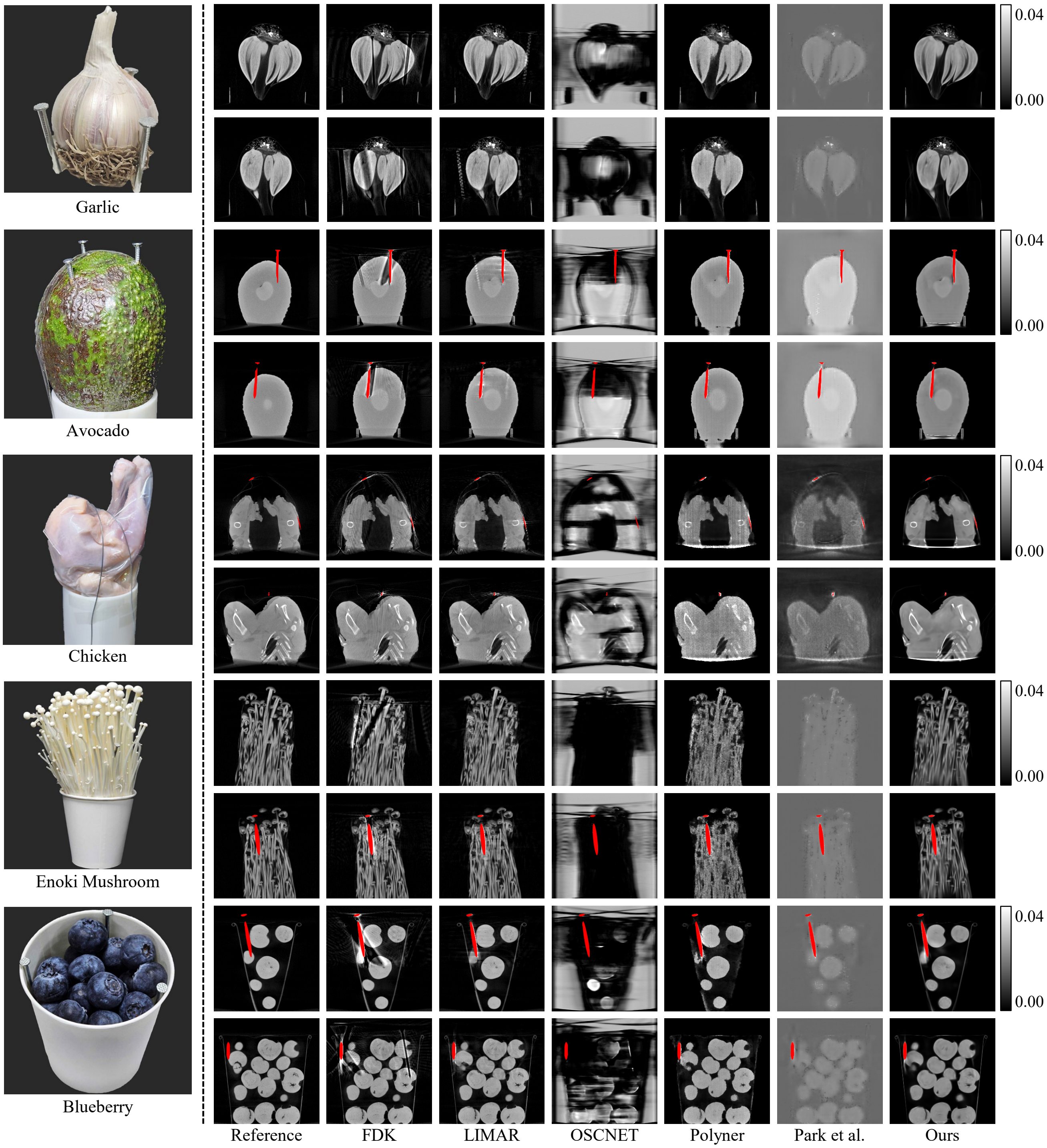


Figure 3. Qualitative comparison on additional slices for real scenes.

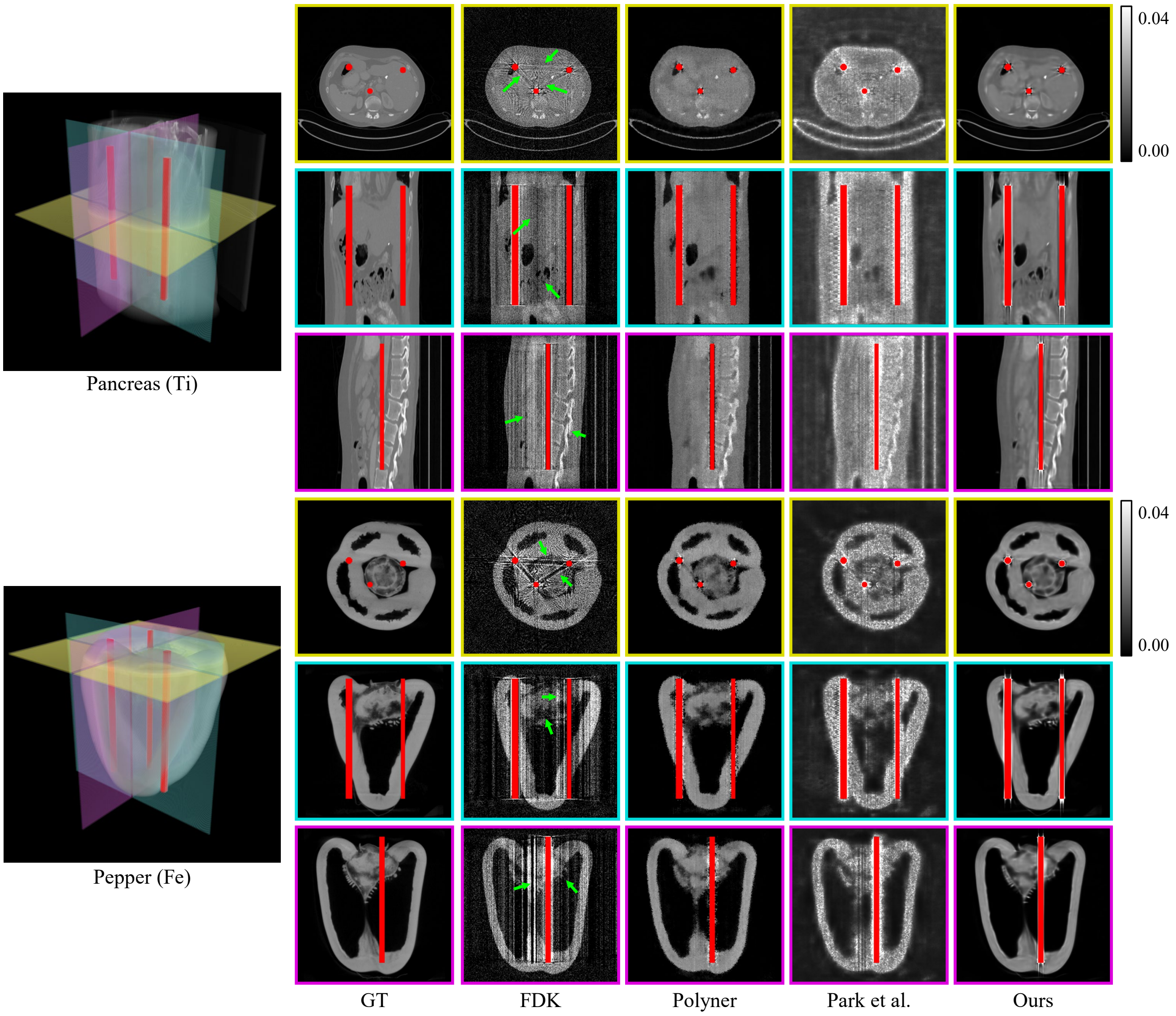


Figure 4. Qualitative comparison on additional synthetic scenes.


\clearpage 
\pagestyle{plain} 
\end{document}